\documentclass[letterpaper, 10 pt, conference]{ieeeconf}
\usepackage{graphics} 
\usepackage{epsfig} 
\usepackage{multirow}
\usepackage{rotating}
\usepackage{colortbl}
\usepackage[table, dvipsnames]{xcolor}
\usepackage{makecell}
\usepackage{amssymb}
\usepackage{tabularx,booktabs}
\usepackage{mathtools}
\usepackage{graphicx}
\usepackage{caption}
\usepackage{subcaption}
\usepackage{amsmath} 
\usepackage{bbm}
\usepackage{tikz}
\usepackage{float}
\usetikzlibrary{positioning}
\usepackage{forest,array}
\usetikzlibrary{calc,shapes,shapes.arrows,arrows,trees,shadows,backgrounds,positioning}
\usepackage{multicol}
\usepackage{balance}

\forestset{
  nodeStyle/.style={
    before typesetting nodes={
      edge=#1,
      for ancestors={
        edge=#1,
        #1,
      },
      #1,
    }
  },
  my edge label/.style={
    edge label={
      node [midway, fill=white, font=\scriptsize] {#1}
    }
  }
}

\IEEEoverridecommandlockouts                              
\title{\LARGE \bf Are We Ready for Robust and Resilient SLAM? \\
A Framework For Quantitative Characterization of SLAM Datasets
}

\author{
	Islam Ali$^{1}$ and Hong Zhang$^{2}$

	\thanks{$^{1}$Islam Ali is with the Department of Computing Science, University of Alberta, Edmonton, AB T6G 2R3, Canada {\tt\small iaali@ualberta.ca}}%

	\thanks{$^{2}$Hong Zhang is with the Department of Electronic and Electrical Engineering, SUSTech, China {\tt\small hzhang@sustech.edu.cn}}%
}
\begin{document}

\maketitle
\thispagestyle{empty}
\pagestyle{empty}

\begin{abstract}
Reliability of SLAM systems is considered one of the critical requirements in modern autonomous systems. This directed the efforts to developing many state-of-the-art systems, creating challenging datasets, and introducing rigorous metrics to measure SLAM performance. However, the link between datasets and performance in the robustness/resilience context has rarely been explored. In order to fill this void, characterization of the operating conditions of SLAM systems is essential in order to provide an environment for quantitative measurement of robustness and resilience. In this paper, we argue that for proper evaluation of SLAM performance, the characterization of SLAM datasets serves as a critical first step. The study starts by reviewing previous efforts for quantitative characterization of SLAM datasets. Then, the problem of perturbation characterization is discussed and the linkage to SLAM robustness/resilience is established. After that, we propose a novel, generic and extendable framework for quantitative analysis and comparison of SLAM datasets. Additionally, a description of different characterization parameters is provided. Finally, we demonstrate the application of our framework by presenting the characterization results of three SLAM datasets: KITTI, EuroC-MAV, and TUM-VI highlighting the level of insights achieved by the proposed framework.

\end{abstract}
\section{Introduction}
The last few decades have witnessed a number of advancements in the field of Simultaneous Localization and Mapping (SLAM). This has been manifested in the introduction of a number of SLAM solutions targeting accuracy and efficiency such as: RTAB-Map \cite{labbe2019rtab}, ORB-SLAM 1,2, and 3 \cite{mur2015orb}\cite{mur2017orb}\cite{campos2020orb}, and VINS-Mono \cite{qin2018vins}, among many others. However, another important consideration of performance, robustness/resilience, has rarely been formally addressed or measured due to the lack of a rigorous definition for it in the SLAM literature. With the increasing need for reliable SLAM solutions in a wide range of critical applications, such as autonomous driving, search and rescue mission, social robotics etc., one may ask whether currently available datasets are able to properly test robustness and resilience of a SLAM system, and whether they can provide us with enough confidence in the system performance both in a challenging situation outside of its tested operating range, and for operating for an extended period of time. In order to measure robustness/resilience in SLAM, a proper definition for them must be achieved first by, for example, exploring how they are defined in other disciplines in science and engineering. For instance, biological systems \cite{stelling2004robustness}\cite{felix2008robustness}, systems engineering \cite{sussman2007building}, and control engineering \cite{steinbuch1991robustness}\cite{stoten1990robustness} define robustness to be the ability of the system to maintain performance under measured perturbations. On the other hand, psychology and ecology \cite{herrman2011resilience} \cite{martin2011resilience}, mechanical and physical robotic systems \cite{negrello2019benchmarking} \cite{koos2013fast}, and system engineering \cite{wied2020conceptualizing} \cite{carpenter2001metaphor} define resilience to be the convergence of a system after divergence while operating outside of its nominal perturbation limits. Perturbations are defined to be any external conditions causing the system to deviate from its equilibrium state. In SLAM, perturbations are usually implicitly contained in benchmark datasets with different sensor measurements and used as inputs to SLAM algorithms. Quantitative characterization of those perturbations leads to the identification of the operating ranges/conditions of SLAM systems and thus, provides a measurement for robustness and resilience of SLAM.

\begin{figure}[t!]
\centering
\includegraphics[width=\columnwidth]{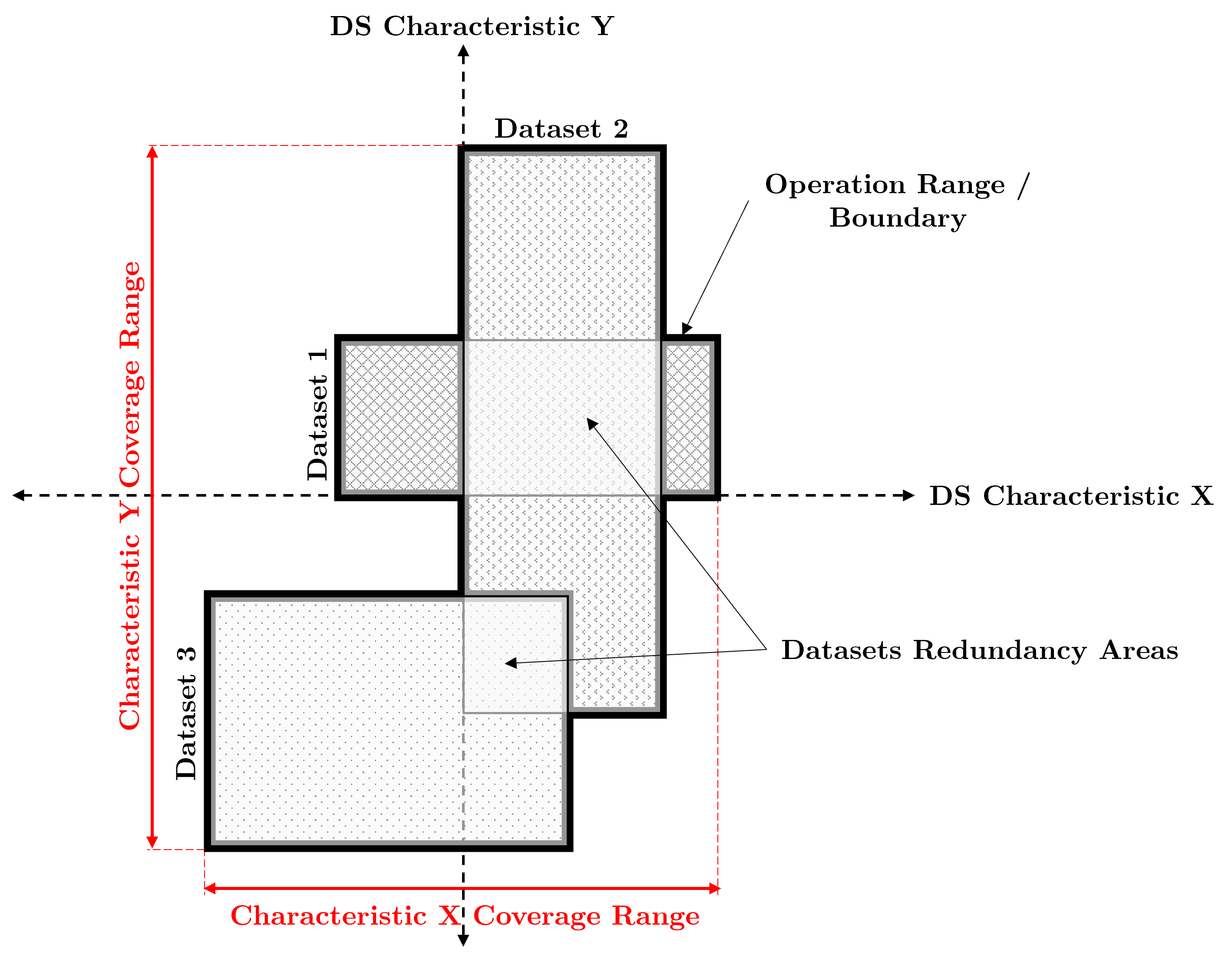}
\caption{An illustration of how the characterization of datasets defines the operating range of SLAM}
\label{fig:operating_range_fig}
\end{figure}

Subsequently, there is an undeniable need to systematically evaluate and compare SLAM datasets based on specified quantifiable metrics. A quantitative analysis of different characteristics present in datasets will not only help identify the suitability of a certain dataset to evaluate robustness or resilience, but also will help in standardizing the process of evaluation of SLAM systems against specifications in general. Fig. \ref{fig:operating_range_fig} illustrates an example of how the characteristics of datasets define the operating ranges of a SLAM system and the limits in which the performance is evaluated and guaranteed.

In order to meet the above need, we introduce a generic and extendable framework for automatic characterization and analysis of SLAM datasets. The framework is designed to provide an efficient way for extension to additional sensors, characteristics, or datasets with limited development efforts. The current version supports visual-inertial SLAM datasets due to their popularity in the SLAM literature.

\begin{figure*}[!tp]
\centering
\includegraphics[width=\textwidth]{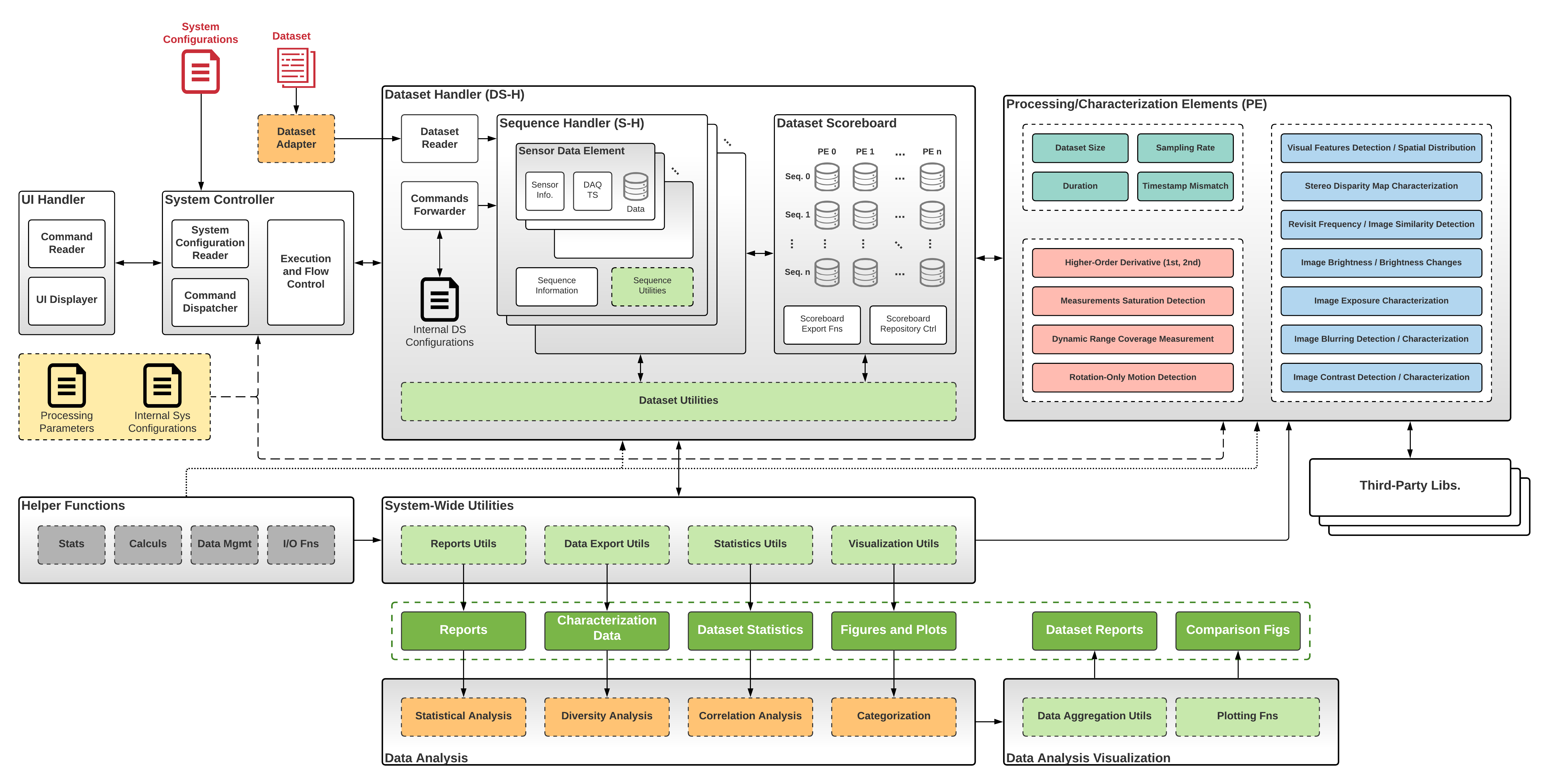}
\caption{A detailed block diagram of the system, illustrating different modules, internal components and the flow of both commands and data throughout the framework}
\label{fig:abstract_BD_detailed}
\end{figure*}

\section{Previous Work}
Developing a SLAM system typically focuses on two aspects: the design of the algorithm itself, and the methodology used for testing and evaluation. The former aspect has received a lot of attention from the SLAM community and led to the development of many advanced and complex systems with different characteristics, sensors support, and architectures. On the other hand, the evaluation of SLAM has focused on the quantification of the localization and mapping quality \cite{fornasier2021vinseval}, and on the introduction of benchmarks and datasets to be used for off-line testing and evaluation. The characterization of the introduced benchmarks/datasets, and the comparison between them has been mostly done qualitatively \cite{liu2021simultaneous}. In this work, we direct the attention of the SLAM community to the importance of the characterization of the datasets themselves rather than the algorithms with the purpose of defining the operating conditions of systems. We provide a measurement of robustness and an evaluation procedure for resilience.  

Quantitative comparison of datasets is typically conducted in the context of introducing a new SLAM algorithm/system with the purpose of justifying the evaluation methodology selected, or introducing a new SLAM dataset to illustrate and elaborate on the differences between the new proposed dataset, and the previously available ones \cite{saeedi2019characterizing}. In the former context, datasets are qualitatively compared with respect to publicly available information. While in the latter context, they are usually compared based on a single metric to provide evidence of how the new dataset is superior compared to its peers. Lately, the focus has started to direct towards SLAM datasets as a discipline of study. For instance, in \cite{liu2021simultaneous}, an extended number of SLAM datasets were reviewed and compared based on a number of qualitative metrics. The paper provided a good overview of the landscape and paved the way to many complementary studies. 

On the other hand, quantitative comparison of datasets can provide some interesting and useful metrics for datasets characterization. However, the generalization of these metrics and the aggregation of them under a single framework for datasets characterization have rarely been attempted, although studies exist on narrow aspects of dataset characteristics. For instance, in \cite{saeedi2019characterizing}, the difficulty of a sequence of robot sensor measurements has been measured independently from the execution of any SLAM system using the Wasserstein distance, which is a statistical method to measure the distance between random variables. By treating each pose as a random variable, the metric can be computed, and thus, is used to measure the level of difficulty of the given sequence. Additionally, the aggressiveness of the robot motion was measured in \cite{Delmerico19icra} using the magnitude of optical flow. Finally, the motion composition and its variability among system axes were measured using principal component analysis (PCA) in \cite{tartanair2020iros} in the analysis of the new challenging TartanAir dataset. A study of the relation between the dataset properties and the SLAM performance was presented in \cite{ye2019characterizing}. In this study, qualitative characteristics of datasets were used as categorical features to build a decision tree to characterize the difficulty of a dataset. Then, the study was extended to explore the relationship between the SLAM performance and the categorical properties. These efforts were directed towards the introduction of characterization metrics, rather than a framework for dataset characterization. The extendibility of our work to other sensors, metrics, and datasets will be evident after the description of the proposed framework.

A closely related line of work has been presented in \cite{andreopoulos2011sensor} and \cite{tsotsos2019does}, and it focuses on the conditions in which a dataset was captured, the parameters of the sensors used for the acquisition, and their impact on computer vision algorithms. In \cite{andreopoulos2011sensor}, the changes in shutter speed, and sensor gain under changing illumination conditions have been discussed. While in \cite{tsotsos2019does}, it was proven that computer vision algorithm performance is experiencing outcome imbalance w.r.t. camera parameters such as ISO, sensor gain, and shutter speed. The results were validated on two different visual datasets. The main difference between this line of work and ours is that, those two studies focus on the conditions in which the data was acquired, while our work focuses on the data after the acquisition process to retrofit available SLAM datasets. 

\section{A Framework for Quantitative Characterization of SLAM Datasets}
In this section we provide a detailed description of the proposed system that ensures the extendability to support additional sensors, datasets, or characterization metrics. 

\subsection{Framework Architecture}
The framework is divided into a number of sub-modules that either work on-line (for data characterization) or off-line (for data analysis and visualization). The system includes a number of built-in configuration files to control different tunable hyperparameters present in the system. In Fig. \ref{fig:abstract_BD_detailed}, a block diagram of all sub-modules of the framework and their interactions are provided in detail.
\subsubsection{Dataset Adaptor}
Datasets differ in terms of the way the data is organized, which requires unification in order for the system to operate seamlessly. Thus, adaptation of datasets to match a certain format is done off-line before the start of the characterization process. This block changes based on the selected dataset and generates a dataset description file which is used by the framework for dataset reading. 
\subsubsection{Dataset Handler}
This module consists of a database that holds sequences, and their enclosed sensor data. The acquisition timestamps are saved alongside the data for synchronization purposes. Additionally, it includes a scoreboard to store the characterization results once calculated from processing elements. Utility functions are also available to facilitate the process of data exchange with other modules.
\subsubsection{Processing Elements Handler}
This module consists of a number of sample/sequence-level processing engines which perform the characterization on dataset data. The characterization results are propagated back to the dataset handler for recording in the dataset scoreboard.
\subsubsection{Dataset Scoreboard}
The dataset scoreboard is a 2-D vector of smaller scoreboards/databases, where each element represents the characterization results of a sequence when run through one of the processing elements.
\subsubsection{UI Handler and System Controller}
The main responsibility of this sub-module is to handle the communication with the user, and control the flow of actions and data among system modules and components. 
\subsubsection{Help Function and System-Wide Utilities}
A number of system utilities are available such as: statistical utilities, calculus utilities, and data conversion/manipulation functions. Additionally, a number of import/export functions are provided to facilitate the later phase of results visualization.
\subsubsection{Data Analysis and Post-Processing}
This module is considered an off-line parallel sub-system, where the characterization results are analyized and visualized after exporting.

\subsection{Characterization Stages}
The characterization process is conducted on three different levels: Sample-level, Sequence-level, and Dataset-level. The system starts by applying different processing elements on applicable dataset samples in each sequence. Then, statistical analysis of the extracted sample-level data is conducted. The objective of this stage is to provide insights on the similarities and the level of diversity of sequences within the same dataset. Finally, statistical analysis of the aggregated data of the whole dataset is done with the objective of comparing datasets and providing rigorous measurement of their implied operating conditions.

\begin{table*}[]
\caption{A Concise Summary of Datasets' Characterization Parameters}
\label{tab:char_table}
\centering
\renewcommand{\arraystretch}{1.3}
\resizebox{\textwidth}{!}{
\begin{tabular}{@{}l|c|ccc|l@{}}
\toprule
\multicolumn{1}{c|}{\textbf{Parameter}} & \textbf{Unit} & \textbf{SM-L} & \textbf{SQ-L} & \textbf{DS-L} & \multicolumn{1}{c}{\textbf{Definition}}                      \\ \bottomrule
\textbf{General Parameters} &  &  &  &  &  \\
$\quad$Measurements Size  & Samples &  & \checkmark & \checkmark & Total number of measured samples / sensor \\
$\quad$Total Duration & Sec. & & \checkmark & \checkmark & $t_{total} = \sum_{i=2}^{N} (t_{i} - t_{i-1})$ \\
$\quad$Sampling Time & Sec. & \checkmark & \checkmark & \checkmark &  $t_{sensor} = (\sum_{i=2}^{N} (t_{i} - t_{i-1})) /N$\\
$\quad$Timestamps Mismatch & Sec. & \checkmark  & \checkmark  & \checkmark  & Timestamps difference between corresponding sensors' samples \\ 
\bottomrule
\textbf{Higher-Order Derivatives} \cite{schot1978jerk} &  &  &  &  &  \\
$\quad$Jerk ($j$) & $m/sec^3$ & \checkmark & \checkmark & \checkmark & $1^{st}$-order time-derivative of acceleration data $[A_x, A_y, A_z]^T$. \\
$\quad$Snap ($S$) & $m/sec^4$ & \checkmark & \checkmark & \checkmark &  $2^{nd}$-order time-derivative of acceleration data $[A_x, A_y, A_z]^T$.\\
$\quad$Angular acc. ($\alpha$) & \textdegree$/sec^2$ & \checkmark & \checkmark & \checkmark & $1^{st}$-order time-derivative of angular velocity data $[G_x, G_y, G_z]^T$. \\
$\quad$Angular jerk ($\varphi$) & \textdegree$/sec^4$ & \checkmark & \checkmark & \checkmark &  $2^{nd}$-order time-derivative of angular velocity data $[G_x, G_y, G_z]^T$. \\
\textbf{Sensor Saturation} &  &  &  &  &  \\
$\quad$Dynamic Range Coverage & $\%$ &  & \checkmark & \checkmark & $ (Max_{sensor}-Min_{sensor}/DR_{sensor}) \%$ \\
$\quad$Dynamic Range Crossing \cite{noureldin2012fundamentals}& $\%$ &  & \checkmark & \checkmark & ($\sum_{i=1}^{n} \mathbbm{1} |x_{(i,sensor)}-DR_{(sensor)}| < DR_{(crossing-ratio)}$) \% \\
\textbf{Rotation-Only Motion} &  &  &  &  &  \\
$\quad$Acceleration Magnitude \cite{noureldin2012fundamentals} & $m/sec^2$ & \checkmark & \checkmark & \checkmark & $f_{mag} = \sqrt{f_{x,b}^2 + f_{y,b}^2 + f_{z,b}^2}$ \\
$\quad$Rotation-Only Samples & $\%$ & \checkmark & \checkmark & \checkmark & ($\sum_{i}^n \mathbbm{1}( f_{magnitude}\geq 9.81\pm10\%)$) \% \\
\bottomrule
\textbf{Image Brightness} &  &  &  &  &  \\
$\quad$Avg. Brightness ($\overline{Br}$) \cite{bezryadin2007brightness}& $DL$ & \checkmark & \checkmark & \checkmark & $\overline{Br} = \sum_{i}^{n} (0.299*R+0.587*G+0.114*B) $ \\
$\quad$Zero-Mean Avg. Brightness Derivative ($\beta_{Br}$) & $DL$ &  & \checkmark & \checkmark & $\beta_{Br} = dBr/dt - \mu_{Br}$ \\
$\quad$Ratio of Thresholding ($\beta_{Br}$) & $\%$ &  & \checkmark & \checkmark & $(\sum_{i}^{n} \mathbbm{1}(|\beta_{Br}| < \sigma_{T}))\%,\quad \sigma_{T} \in \{\sigma_{Br}, 2\sigma_{Br}, 3\sigma_{Br}\}$ \\
\textbf{Image Exposure} &  &  &  &  &  \\
$\quad$Trimmed Image Mean ($\mu_{\alpha}$) \cite{rice2006mathematical} & $DL$ & \checkmark & \checkmark & \checkmark & $\mu_{\alpha} = (1)/(n-2[n\alpha])* \sum_{i=n-[n\alpha]}^{[n\alpha+1]} I_i$ \\
$\quad$Trimmed Image Skewness ($S_{\alpha}$) \cite{rice2006mathematical} & $DL$ & \checkmark & \checkmark & \checkmark & $S_{\alpha} = (\sum_{i=n-[n\alpha]}^{[n\alpha+1]} (I_i - \mu_{\alpha})^2)/((n-2[n\alpha]-1)*\sigma_{\alpha}^3)$ \\
$\quad$Exposure Zone \cite{gibson2014exposure} & $DL$ & \checkmark & \checkmark & \checkmark & Detection of black, white, under-, over-, or properly exposed images\\
\textbf{Image Contrast} \cite{peli1990contrast} &  &  &  &  &  \\
$\quad$Contrast Ratio ($C_{CR}$) & $DL$ & \checkmark & \checkmark & \checkmark & $C_{CR} = (L_{target}/L_{background}) *100$ \\
$\quad$Weber Contrast ($C_{W}$) & $DL$ & \checkmark & \checkmark & \checkmark & $C_{W} = ((L_{target}- L_{background})/L_{background}) * 100$ \\
$\quad$Michelson Contrast ($C_{M}$) & $DL$ & \checkmark & \checkmark & \checkmark &  $C_{M} = (L_{max} - L_{min})/(L_{max} + L_{min})$\\
$\quad$RMS Contrast ($C_{RMS}$) & $DL$ & \checkmark & \checkmark & \checkmark & $C_{RMS} = \sqrt{1/n \sum_{i=1}^{n} (l_i - \bar{l})^{2}}$ \\
\textbf{Image Blurring} \cite{pertuz2013analysis}&  &  &  &  &  \\
$\quad$Blurring Score ($\sigma_{\nabla^2}$)& $DL$ & \checkmark & \checkmark & \checkmark & Variance of the laplacian $(\sigma^2_{\nabla^2}) = variance(\partial^2 I/\partial x^2 * \partial^2 I/\partial y^2)$ \\
$\quad$Blurring Percentage/Image & $\%$ & \checkmark & \checkmark & \checkmark & $(\sum_i^{n} \mathbbm{1} \sigma^2_{\nabla^2,sub-img} > Blurring_{Threshold})\% $ \\
$\quad$Blurred Images Percentage & $\%$ &  & \checkmark & \checkmark &$(\sum_i^{n} \mathbbm{1} \sigma^2_{\nabla^2,img} > Blurring_{Threshold})\% $ \\
\textbf{Detectable V-Features(SIFT\cite{lowe2004distinctive}, ORB\cite{rublee2011orb}, FAST\cite{rosten2006machine})} &  &  &  &  &  \\
$\quad$Avg. \# Feature/sub-image ($F_{avg}$) & $Features$ & \checkmark & \checkmark & \checkmark &  $F_{avg} = (F_{total})/(\lceil{I_H/b_{dim}}\rceil * \lceil{I_W/b_{dim}}\rceil)$\\
$\quad$Avg. Spatial distribution Ratio ($F_{dist-avg}$) & $\%$ & \checkmark & \checkmark & \checkmark &  $F_{dist-avg} = (\sum_{i=1}^{n} \mathbbm{1} (F_i >= F_{avg}))/(b_{total}) \%$\\ 
$\quad$Abs. Spatial distribution Ratio ($F_{dist-abs}$) & $\%$ & \checkmark & \checkmark & \checkmark &  $F_{dist-abs} = (\sum_{i=1}^{n} \mathbbm{1} (F_i >= 1))/(b_{total})\%$\\ 
\textbf{Image Disparity} &  &  &  &  &  \\
$\quad$Avg. Disp.(StereoBM) ($\mu_{D,BM}$) \cite{konolige2010projected} & $DL$ & \checkmark & \checkmark & \checkmark &  Average of Disparity Map using StereoBM Method\\
$\quad$Std. Dev. Disp.(StereoBM) ($\sigma_{D,BM}$) \cite{konolige2010projected} & $DL$ & \checkmark & \checkmark & \checkmark &  Standard dev. of Disparity Map using StereoBM Method\\
$\quad$Avg. Disp.(StereoSGBM) ($\mu_{D,SGBM}$) \cite{hirschmuller2007stereo} & $DL$ & \checkmark & \checkmark & \checkmark &  Average of Disparity Map using StereoSGBM Method\\
$\quad$Std. Dev. Disp.(StereoSGBM) ($\sigma_{D,SGBM}$) \cite{hirschmuller2007stereo} & $DL$ & \checkmark & \checkmark & \checkmark &  Standard dev. of Disparity Map using StereoSGBM Method\\
\textbf{Image Similarity} \cite{GalvezTRO12} &  &  &  &  &  \\
$\quad$DBoW2 Similarity Score & $DL$ & \checkmark & \checkmark & \checkmark & DBoW2 score to closet match in the same sequence  \\
$\quad$Distance to Closest Match & $Frames$ & \checkmark & \checkmark & \checkmark & The proximity distance between an image and its closest match in a sequence  \\
\bottomrule
\multicolumn{6}{l}{* SM-L, SQ-L, and DS-L refers to sample-, sequence-, and dataset level characterization}\\
\multicolumn{6}{l}{* DR refers to inertial sensor's dynamic range}\\
\multicolumn{6}{l}{* DL refers to dimensionless quantity}\\
\end{tabular}
}
\end{table*}

\subsection{Characterization Assumptions}
The framework uses the default values of the configuration parameters of any underlying engines such as feature extraction, disparity calculations, etc. Tunable parameters such as threshold values can be controlled from within the framework. Both default and tunable parameters used for the characterization process are kept exactly the same to ensure fairness of comparison.
\subsection{SLAM Dataset Characterization Parameters}
A number of characterization parameters were selected for characterizing SLAM datasets. The characterization parameters are categorized into three different groups, according to the aforementioned characterization levels of the framework. The different characterization parameters are summarized in Table \ref{tab:char_table}.
\subsubsection{General Characterization Parameters}
The parameters of this category are concerned with non-sensory characteristics of a dataset. Occasionally,  these parameters are qualitatively reported but not systematically measured. 
\subsubsection{Inertial Characterization Parameters}
Inertial Measurement Units (IMUs) are one of the most popular SLAM sensors due to their competitive price and the associated acceptable performance. Motion profiles can be deduced from the characterization of the IMU data, and thus, systematic characterization is supported in the proposed framework.
\subsubsection{Visual Characterization Parameters}
RGB cameras are considered a fundamental sensor in SLAM due to the richness of information one can extract from them. Subsequently, many environment and data capturing characteristics can be deduced from the quantitative analysis of images. In the proposed framework, the concept of sub-images was used while measuring image blurring and spatial distribution of visual features as such; an image is divided into sub-images, and measurement is conducted on each sub-image on its own. Then, the measurements are aggregated to describe the whole image. The concept of sub-images is useful in determining local events or anomalies in the images, which are present in challenging scenes. 
 \begin{table*}[htbp]
\caption{Dataset Characterization Results' Analysis Metrics}
\centering
\label{tab:analysis_table}
\begin{tabular}{@{}l|l|l@{}}
\toprule
\textbf{Analysis Metric}               & \textbf{Formula} & \textbf{Parameters} \\ \toprule
Statistical Analysis                   &
$\mu$, $mid$, $\sigma$, $\sigma^2$, $S$&
\makecell[l]{
$N$ total number of samples\\
$x$ input vector
}\\  \midrule

Entropy (H)                            & 
$H = -\sum_{i=0}^k \; p_i \; log_2 \; (p_i)$ & 
\makecell[l]{
$k$ no. of unique values\\
$p_i$ prob. associated
}\\ \midrule

Simpson Diversity Index (SDI)          & 
$SDI = 1 \;-\; \frac{\sum_{i=1}^k n_i (n_i -1)}{N(N-1)}$ &
\makecell[l]{
$N$ total number of samples \\
$k$ no. of unique values\\
$n_i$ no. of samples for a unique value
}\\ \midrule

Pearson Correlation Coefficient (PPMC) &  
$PPMC(x,y) = \frac{N \sum(x.y) - \sum(x)\sum(y)}{\sqrt{[N \sum(x^2) - (\sum(x)^2)].[N \sum(y^2) - (\sum(y)^2)]}}$ &
\makecell[l]{
$N$ total number of samples \\ 
$x$ first characterization metric vector \\
$y$ second characterization metric vector \\
}\\ \bottomrule
\end{tabular}
\end{table*}
\section{Experimental Results and Discussion}
To systematically analyze and compare measured characteristics, dataset characterization techniques from the field of data science \cite{pearson2011exploring} are used. The analysis includes: \textit{Statistical Analysis}, \textit{Diversity And Interestingness Analysis} where Shannon Entropy $(H)$ and Simpson Diversity Index $(SDI)$ are calculated, and \textit{Correlation Analysis} where Pearson Correlation Coefficient $(PPMC)$ is measured between the characterization metrics and performance metrics of SLAM algorithms. The analysis metrics are provided in Table. \ref{tab:analysis_table}.

The framework has been used to characterize three datasets which, together, are considered a classical standard for benchmarking SLAM. These datasets are KITTI Odometry \cite{geiger2012we}, EuroC MAV \cite{burri2016euroc}, and TUM VI \cite{schubert2018vidataset}. Due to the extendability and generality of the framework, support for additional characterization parameters, sensors, or datasets is possible with minimal development efforts. 

The importance of the characterization becomes evident when one takes a deeper look at the results as many insights can be extracted and measured. The following points highlight some of the benefits for the SLAM research and development community. The framework provides a foundation for a complete ecosystem for a systematic and a reliable methodology for SLAM algorithms development, testing, and performance evaluation.

\begin{table}[]
\centering
\caption{Statistical analysis of Entropy Results}
\label{tab:H_table}
\begin{tabular}{@{}lccc@{}}
\cmidrule(l){2-4}
                                & \multicolumn{3}{c}{mean$\pm$std. dev}        \\ \midrule
\multicolumn{1}{l|}{}           & KITTI & EURO-C & TUM-VI \\ \midrule
\multicolumn{1}{l|}{General Metrics}        &2.0$\pm$(1.67) & 1.73$\pm$(0.86) & 3.81$\pm$(2.05) \\
\multicolumn{1}{l|}{Inertial Metrics}       &0.0$\pm$(0.0) & 6.42$\pm$(2.83) & 10.12$\pm$(4.39) \\
\multicolumn{1}{l|}{Visual Metrics}         &6.17$\pm$(4.33) & 5.95$\pm$(4.03) & 6.86$\pm$(4.46) \\ \bottomrule
\end{tabular}
\end{table}

\begin{table}[]
\centering
\caption{Statistical analysis of SDI Results}
\label{tab:SDI_table}
\begin{tabular}{@{}lccc@{}}
\cmidrule(l){2-4}
                                & \multicolumn{3}{c}{mean$\pm$std. dev}        \\ \midrule
\multicolumn{1}{l|}{}           & KITTI & EURO-C & TUM-VI \\ \midrule
\multicolumn{1}{l|}{General Metrics}        &0.59$\pm$(0.48) & 0.78$\pm$(0.32) & 0.96$\pm$(0.06) \\
\multicolumn{1}{l|}{Inertial Metrics}       &0.0$\pm$(0.0) & 1.0$\pm$(0.0) & 1.0$\pm$(0.0) \\
\multicolumn{1}{l|}{Visual Metrics}         &0.74$\pm$(0.42) & 0.8$\pm$(0.35) & 0.85$\pm$(0.32) \\ \bottomrule
\end{tabular}
\end{table}

\subsection{Dataset Diversity and Interestingness}
\begin{figure}[!tp]
     \centering
     \begin{subfigure}[b]{0.31\columnwidth}
         \centering
         \includegraphics[width=\columnwidth]{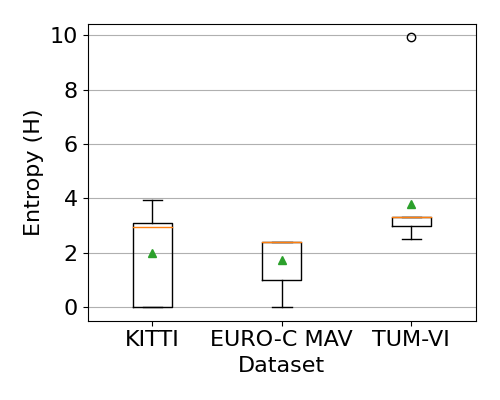}
         \caption{}
     \end{subfigure}
     \hfill     
     \begin{subfigure}[b]{0.31\columnwidth}
         \centering
         \includegraphics[width=\columnwidth]{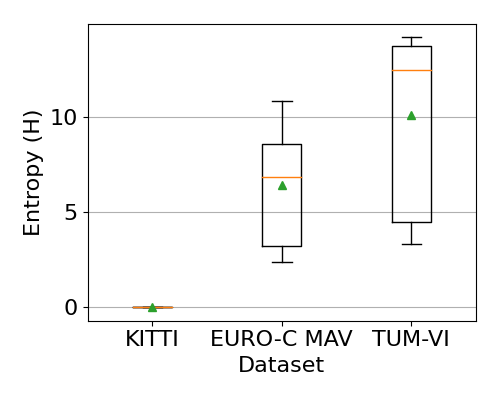}
         \caption{}
     \end{subfigure}      
	 \hfill	 
     \begin{subfigure}[b]{0.31\columnwidth}
         \centering
         \includegraphics[width=\columnwidth]{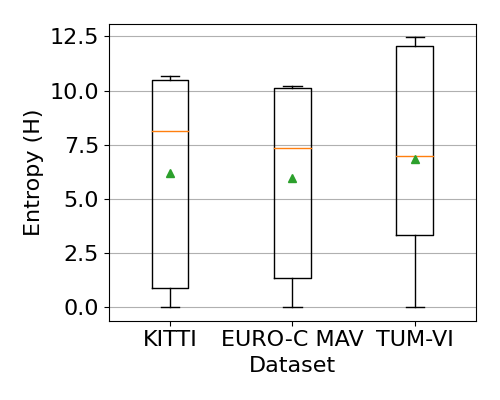}
         \caption{}
     \end{subfigure}
     \hfill
        \caption{Entropy analysis of (a)general, (b)inertial, and (c)visual characterization results respectively}
        \label{fig:H_figs}
\end{figure}
\begin{figure}[!tp]
     \centering
     \begin{subfigure}[b]{0.31\columnwidth}
         \centering
         \includegraphics[width=\columnwidth]{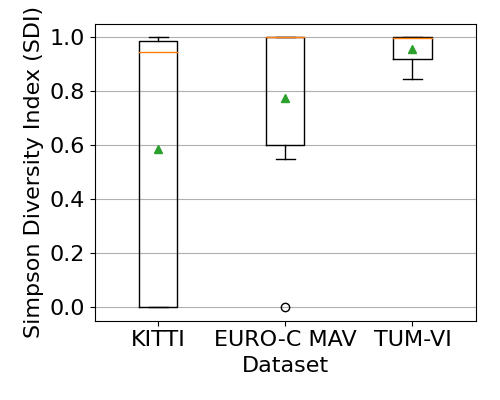}
         \caption{}
     \end{subfigure}
     \hfill     
     \begin{subfigure}[b]{0.31\columnwidth}
         \centering
         \includegraphics[width=\columnwidth]{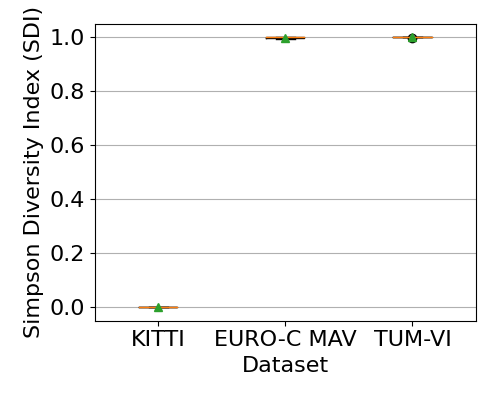}
         \caption{}
     \end{subfigure}      
	 \hfill	 
     \begin{subfigure}[b]{0.31\columnwidth}
         \centering
         \includegraphics[width=\columnwidth]{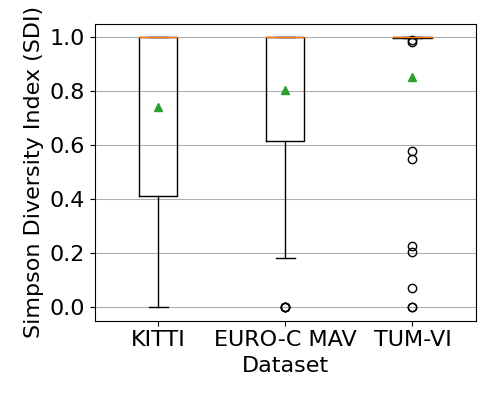}
         \caption{}
     \end{subfigure}
     \hfill
        \caption{Simpson's Diversity Index (SDI) analysis of (a)general, (b)inertial, and (c)visual characterization results respectively}
        \label{fig:SDI_figs}
\end{figure}

In Figure \ref{fig:H_figs} and Figure \ref{fig:SDI_figs}, entropy and SDI analysis results are presented. It can be shown that although the three datasets are very diverse, the amount of information measured by the entropy and the SDI varies due to the nature of the dataset formulation. Moreover, one can observe that the TUM-VI dataset is superior in diversity when compared to EURO-C MAV and KITTI in terms of both general and visual characterization. However, the diversity of TUM-VI and EURO-C MAV is almost the same when it comes to inertial characteristics.

\begin{figure}[!tp]
     \centering
     \begin{subfigure}[b]{0.45\columnwidth}
         \centering
         \includegraphics[width=\columnwidth]{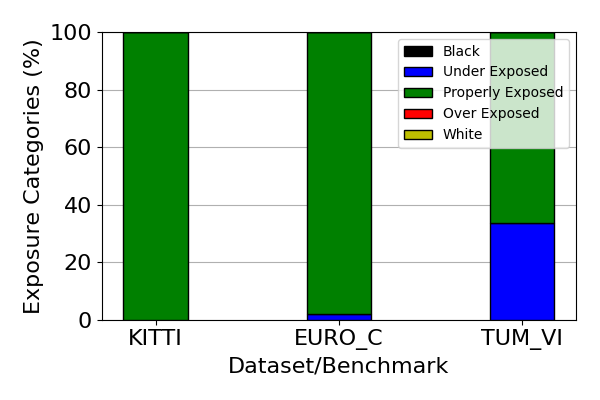}
         \caption{Exposure Zones}
         \label{fig:exp_zone}
     \end{subfigure}
     \hfill     
     \begin{subfigure}[b]{0.45\columnwidth}
         \centering
         \includegraphics[width=\columnwidth]{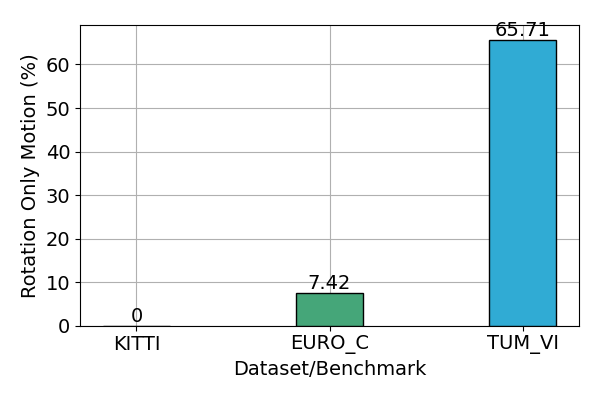}
         \caption{Rotation-only Motion}
         \label{fig:rot_only}
     \end{subfigure}      
     \hfill
        \caption{Example dataset anomalies detected using the analysis of the characterization results}
        
\end{figure}
\subsection{Dataset Anomalies} 
Dataset anomalies can be detected using the proposed framework. For instance, sensor timestamp mismatch has been detected in the TUM-VI dataset, and this imposes a data integrity issue for SLAM. Additionally, over-exposed and under-exposed images have also been detected and localized in both EuroC and TUM-VI datasets as shown in Figure \ref{fig:exp_zone}. Moreover, rotation-only motion profiles have been detected in EuroC and TUM-VI as shown in Figure \ref{fig:rot_only}, and this implies a challenge on conventional SLAM solutions. The existence of these anomalies proposes that the TUM-VI dataset is more diverse in terms of existence of data anomalies compared to the other two datasets.

\begin{figure}[!tp]
     \centering
     \begin{subfigure}[b]{0.31\columnwidth}
         \centering
         \includegraphics[width=\columnwidth]{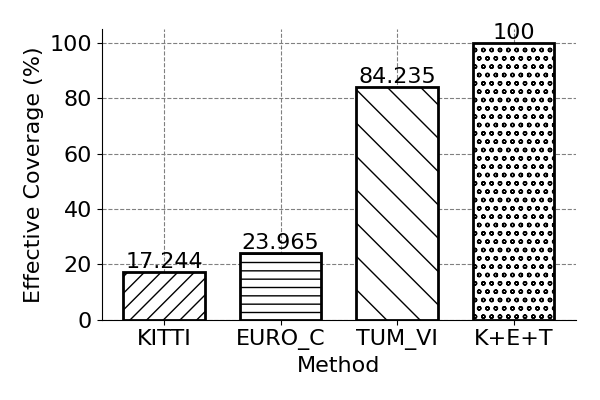}
         \caption{}
     \end{subfigure}
     \hfill     
     \begin{subfigure}[b]{0.31\columnwidth}
         \centering
         \includegraphics[width=\columnwidth]{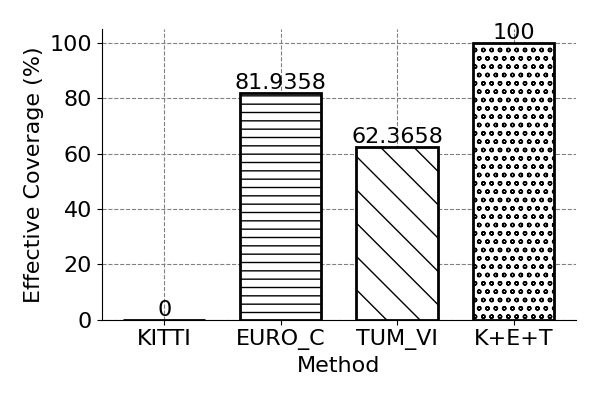}
         \caption{}
     \end{subfigure}      
	 \hfill	 
     \begin{subfigure}[b]{0.31\columnwidth}
         \centering
         \includegraphics[width=\columnwidth]{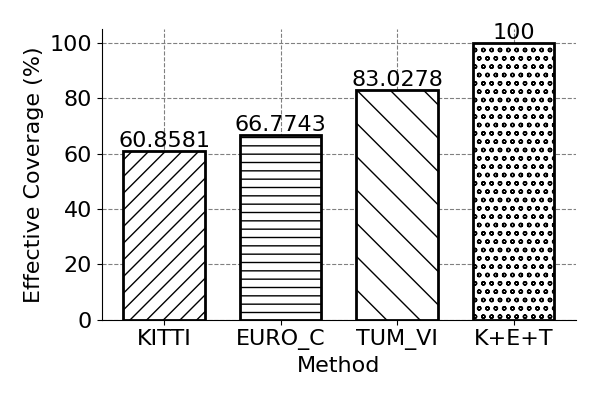}
         \caption{}
     \end{subfigure}
     \hfill
        \caption{Dynamic range coverage percentage of (a)general, (b)inertial, and (c)visual characterization metrics of each dataset compared to their aggregation}
        \label{fig:coverage_fig}
\end{figure}
\subsection{Operating Conditions and SLAM Robustness/Resilience}
The operating conditions and dynamic ranges of a SLAM system can finally be defined quantitatively using the proposed framework. In Figure \ref{fig:coverage_fig}, datasets are compared based on the level of dynamic range coverage achieved compared to the usage of the three of them combined. From the perspective of general and visual characterization metrics, TUM-VI is superior in terms of the achieved coverage. However, the EuroC-MAV dataset achieved a higher coverage when it comes to inertial characterization metrics. Claimed performance of SLAM is now tied to an operating range which is a direct indication of how robust the system is. Subjecting a SLAM systems to perturbation beyond its operating conditions until divergence is also possible and is a measure of the system rating (operating limits) in which a system can survive. With the knowledge of the operating limits, a measurement of resilience can be realized. This can be achieved by subjecting SLAM to conditions beyond the operating limits and measuring the time to convergence and the associated accuracy. 

\begin{figure}[!tp]
\centering
\includegraphics[width=\columnwidth]{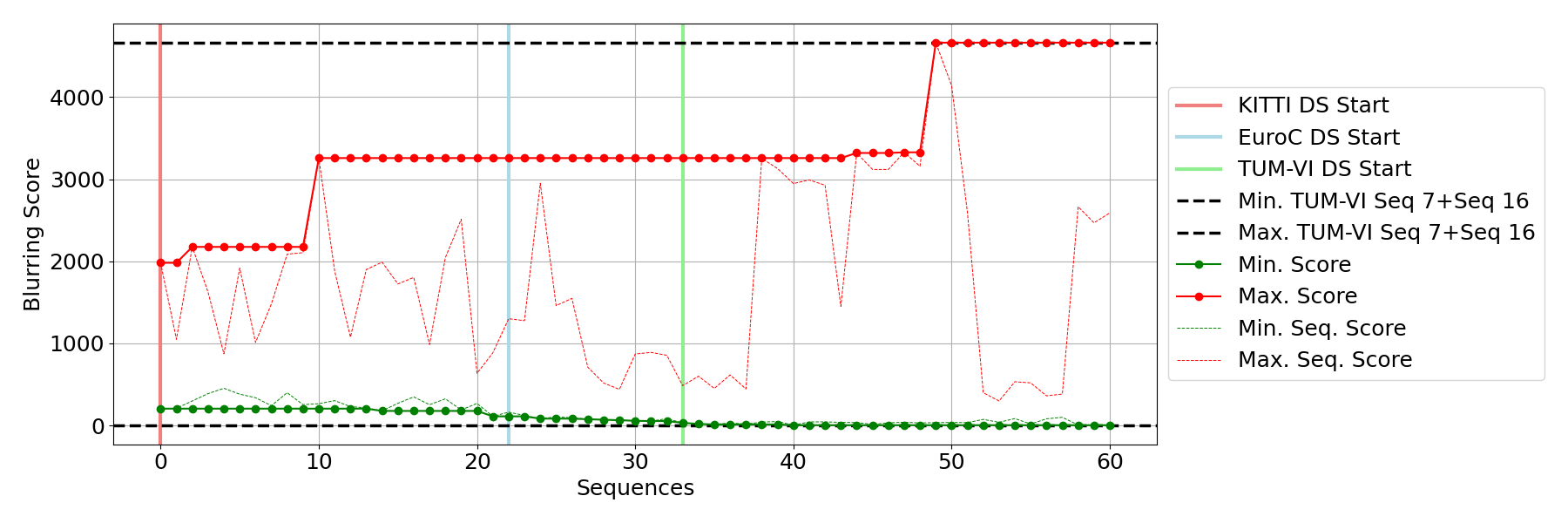}
\caption{Blurring Score Coverage Analysis}
\label{fig:blurring_coverage}
\end{figure}
\subsection{Dataset Redundancy}
Redundancy can either happen within a dataset among its sequences or among different datasets used for the same evaluation process. The detection of either can better guide the design of experiments via the proper selection of datasets or sequence mixes. This can pave the way to a systematic procedure for SLAM evaluation and will act as an entry point for \textit{task-centric} selection of the evaluation and testing methodology and data. Moreover, the novelty of any newly introduced dataset can be assessed in comparison to what we already have via a systematic evaluation/comparison engine. For example, coverage analysis was conducted on the blurring score of images. The results provided in Fig. \ref{fig:blurring_coverage} suggest that only two sequences from the TUM-VI dataset (Seq. 7 and Seq. 16) are sufficient for testing SLAM immunity to blurring as they provides the same coverage levels achieved by testing all the three datasets for blurring. Subsequently, the testing time of SLAM against blurring was reduced from running sixty-one sequences from three different datasets to only running two sequences of the TUM-VI dataset. 

\begin{figure}[tp]
     \centering
         \centering
         \includegraphics[width=\columnwidth]{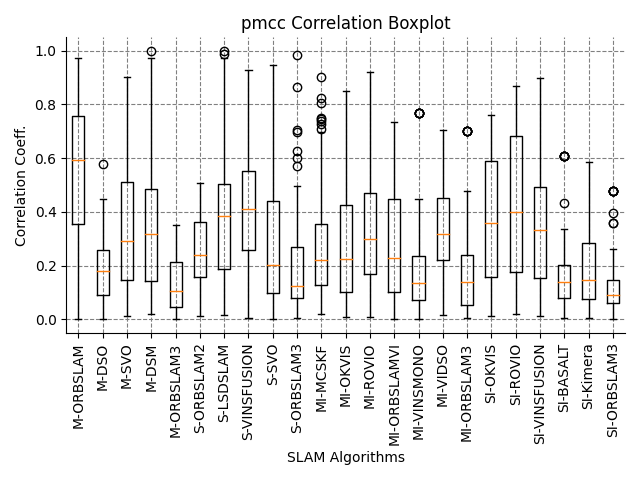}    
        \caption{PMCC Correlation Coefficients between characterization metrics mean for each sequence and the reported ATE of selected SLAM algorithms. (M, S, MI, SI) refer to (Mono, Stereo, Mono-Inertial, Stereo-Inertial) operating modes.}
        \label{fig:corr_results}
\end{figure}

\subsection{Dataset and Performance Correlation Analysis}
The availability of dataset characteristics/features can be used in analyzing the correlation among characteristics or between characteristics and SLAM performance revealing the underlying relations between the two and pin-pointing the causes for system failures, performance degradation, or system sensitivity to input data. Consequently, an important aspect of this work is the illustration of how dataset characterization can lead to predictable SLAM performance. Thus, in Figure \ref{fig:corr_results}, the correlation between the ATE of SLAM and the mean of characterization results is presented. We can observe the existence of highly correlated characterization metrics with ATE, which shows the significance of the selected characterization metrics and their suitability for usage as a dataset descriptor.

\section{Conclusion}
In this work, the problem of quantitative characterization of SLAM datasets is discussed. It has been shown that, quantitative characterization of datasets can provide a measurement for the operating ranges and conditions and serve as a key to measuring robustness and resilience. Moreover, such characterization can provide a systematic methodology for designing experiments where a certain environment condition is exploited. Thus, a novel dataset characterization framework has been introduced and described in detail. As an example, the framework is used to characterize three different datasets with the objective of showing its capabilities in terms of measurements and visualization. The characterization process highlights a number of undiscovered anomalies present in some of the datasets and opens door to a wide range of studies that can be conducted. The link between the characteristics of the data and the algorithm performance can finally be established and can lead to a systematic evaluation methodology for SLAM system research and development. On the other hand, introduction of new datasets can be guided by the framework outcomes in terms of detecting anomalies by providing a measure for diversity, redundancy, and coverage. 
\bibliographystyle{IEEEtran}
\balance
\bibliography{refs}

\end{document}